\documentclass[conference]{IEEEtran}
\IEEEoverridecommandlockouts

\usepackage{cite}
\usepackage{amsmath,amssymb,amsfonts}
\usepackage{algorithmic}
\usepackage{graphicx}
\usepackage{textcomp}
\usepackage{xcolor}
\usepackage{tabularx}
\usepackage{graphics} % for pdf, bitmapped graphics files
\usepackage{epsfig} % for postscript graphics files
\usepackage{mathptmx} % assumes new font selection scheme installed
\usepackage{times} % assumes new font selection scheme installed
\usepackage{amsmath} % assumes amsmath package installed
\usepackage{amssymb}  % assumes amsmath package installed
\usepackage{booktabs} % for \toprule, \midrule, \bottomrule
\usepackage{cite}
\def\BibTeX{{\rm B\kern-.05em{\sc i\kern-.025em b}\kern-.08em
    T\kern-.1667em\lower.7ex\hbox{E}\kern-.125emX}}
\begin{document}

\title{Video Reconstruction using Diffusion-based Image-to-Video Generation with Trajectory Guidance}

\author{\IEEEauthorblockN{Stelio Bompai$^{1}$, Ioannis Kontopoulos$^{1}$, Giannis Spiliopoulos$^{2}$, Dimitris Zissis$^{2}$, Konstantinos Tserpes$^{1}$}
\IEEEauthorblockA{\textit{$^{1}$Department of Electrical and Computer Engineering, National Technical University of Athens, Greece}\\
\textit{$^{2}$Department of Product and Systems Design Engineering, University of the Aegean, Syros, Greece
}}}

\maketitle
\begin{abstract}
This paper addresses the problem of reconstructing missing or dropped 
frames in top-down drone video of autonomous surface vehicles 
performing structured maritime manoeuvres. We propose a pipeline that 
converts raw GPS telemetry and a single reference frame into a 
trajectory-guided video sequence using a pre-trained image-to-video 
diffusion model, requiring no domain-specific fine-tuning. GPS 
coordinates from onboard telemetry logs are projected into image space 
via an equirectangular mapping, producing per-vessel motion cues that 
condition the SG-I2V diffusion model. The generated frames are 
evaluated against ground-truth video using perceptual, temporal and 
trajectory-based metrics, and benchmarked against optical flow 
extrapolation and RIFE interpolation baselines. SG-I2V produces the 
most naturally appearing frames among all methods (BRISQUE 25.52, 
closest to ground-truth 23.64), the most realistic motion magnitude (temporal 
smoothness 1.14 vs.\ ground truth 1.42), and the strongest GPS trajectory adherence (9.31px vs.\ 28.70px for ground-truth, the latter reflecting approximate temporal alignment between footage and GPS logs rather than generation error), 
demonstrating that trajectory-guided diffusion synthesis is a viable 
approach to maritime video reconstruction under challenging low-texture, 
small-object conditions.
\end{abstract}


\section{Introduction}

Video data is a foundation of modern surveillance, environmental monitoring systems, and autonomous navigation. In practice, however, video streams are frequently corrupted or contain missing segments resulting from a diverse set of factors, including network transmission errors and packet loss, hardware and sensor malfunctions~\cite{wang1998error}, adverse environmental conditions such as fog, glare, or precipitation, degradations that are introduced by the platform such as vibration and motion blur in airborne systems, and environmental conditions such as object occlusion or sudden illumination changes~\cite{huang2025systematic}. Reconstructing plausible scenes for these missing segments is a non-trivial problem~\cite{quan2024deep} because, unlike single-image inpainting, video reconstruction must respect both spatial coherence within each frame and temporal consistency across the sequence~\cite{liu2025appearance}.
Traditional approaches to this challenge have largely relied on statistical interpolation, linear or spline-based methods that blend neighboring frames under the implicit assumption of predictable, smooth motion~\cite{dong2023video}. While computationally cheap, these methods break down under fast or non-linear object dynamics and cannot synthesize plausible texture or structure in heavily corrupted regions~\cite{han2024motion}.

Machine learning methods, particularly convolutional and recurrent architectures, have pushed beyond interpolation by learning data-driven motion priors. But they still remain constrained by the training distribution and tend to produce blurry outputs when conditioned on sparse observations alone~\cite{huang2025systematic, liu2025appearance}.
Generative modeling and diffusion-based methods in particular have emerged as a new standard for visual synthesis~\cite{ho2020denoising}. By learning a progressive denoising process over high-dimensional image distributions, diffusion models produce results that are both high quality and diverse, something earlier GAN-based models found difficult to achieve~\cite{dhariwal2021diffusion}. State-of-the-art image-to-video diffusion models extend this capability to the temporal domain in which case, given a single conditioning frame, they generate a temporally coherent sequence by iteratively refining a stochastic video trajectory within learned dynamics priors~\cite{blattmann2023stable, ma2025video}.

An important gap, however, persists when the visual signal alone is insufficient to determine object motion, a common situation in wide-area maritime or aerial footage where objects are small relative to the field of view. In such settings, auxiliary information from secondary sensors or tracking systems can provide the motion knowledge that the image signal lacks. Trajectory data, encoding the time-stamped spatial positions of tracked objects, is one such source. If this information can be projected into a form a generative model can condition on, it becomes possible to steer video synthesis toward physically plausible outcomes.

This paper proposes a pipeline that bridges these two modalities. Given a single observed video frame and a set of object trajectories obtained from a secondary source (i.e., CSV), recording the coordinates of small boats within the scene, we construct per-object arrow overlays by mapping real-world coordinates into image space via a GPS-to-pixel mapping function. The annotated frame is then passed as input to a state-of-the-art image-to-video diffusion model, which generates the subsequent 14 frames conditioned on both the visual content and the embedded trajectory guidance. The reconstructed frames are evaluated against ground-truth video to quantify the fidelity and temporal coherence of the synthesis. Our contributions are: (i) a GPS-to-image projection scheme that converts sparse tracking data into spatial motion cues consumable by off-the-shelf generative models (ii) a complete reconstruction pipeline that requires no model fine-tuning and (iii) a systematic evaluation benchmarking of the proposed pipeline against traditional video frame interpolation and extrapolation methods on real maritime footage, demonstrating the advantages of generative trajectory-guided synthesis under challenging small-object conditions.

\section{Related Work}

\subsection{Statistical and Classical Methods}
The earliest efforts at video frame reconstruction treated the problem as temporal signal interpolation. Optical flow estimation, pioneered by Horn and Schunck~\cite{horn1981determining} and later extended by polynomial-expansion methods such as Farneb\"{a}ck's~\cite{farneback2003two} as implemented in OpenCV library, provided a mechanism for warping existing frames toward predicted positions and remains a component of many modern pipelines. Interpolation methods based on motion compensation refined this idea by explicitly modeling displacement fields at the block level or pixel level, and have been widely deployed in video compression standards~\cite{dong2023video}. 

In their simplest form, however, these methods operate from a single reference frame, warping it forward by a scaled flow field. This compounds errors over time, as without a second anchor frame to guide the synthesis, small flow estimation errors build up with each step, leading to increasingly degraded outputs the further apart the frames are. Furthermore, all approaches that rely purely on warping are fundamentally limited in their ability to generate content that does not exist in the reference frame, and perform poorly when objects become occluded, move significantly, or change in appearance~\cite{han2024motion}.

\subsection{Deep Learning-Based Video Synthesis}
Convolutional neural networks brought a data-driven perspective to video generation, learning to extrapolate future frames from short observed historical frames. Later work incorporated adversarial objectives to sharpen temporal generation~\cite{mathieu2015deep}, and recurrent architectures such as ConvLSTMs enabled models to maintain a compact latent memory of dynamics, improving coherence over longer horizons. 
%I need to change the transformer-based implication here that connects immediately to the RIFE.
More recently, transformer-based video models have scaled to high resolutions and long sequences, leveraging spatiotemporal attention to capture non-local dependencies. Learned interpolation methods such as RIFE~\cite{huang2022real} address the compounding error problem of single-frame warping by conditioning on \textit{both} temporal endpoints, directly estimating bidirectional intermediate flow fields $F_{0\to t}$ and $F_{1\to t}$ and fusing the warped results with a learned blending network. This bilateral anchoring substantially improves stability across the interpolated sequence. As discussed in the previous subsection, this fundamental limitation of approaches based on warping, persists even in more sophisticated deep learning variants, ultimately motivating the shift toward generative approaches.

\subsection{Generative Models and Diffusion-Based Video Generation}
Early generative approaches introduced adversarial objectives to sharpen temporal generation, but such methods often require large domain-specific training sets and are prone to mode collapse under high uncertainty, producing averaged, over-smoothed outputs when conditioned on ambiguous or sparse inputs~\cite{dhariwal2021diffusion}. Denoising diffusion probabilistic models (DDPMs), introduced by Ho et al.~\cite{ho2020denoising} and extended to the conditional setting by numerous subsequent works, reformulate synthesis as iterative noise removal guided by a learned score function. Their application to video began with architectures such as Video Diffusion Models~\cite{ho2022video}, which factorized space-time attention to generate temporally consistent clips. Stable Video Diffusion (SVD)~\cite{blattmann2023stable} demonstrated that a diffusion model conditioned on a single image and pre-trained on large-scale video data can generalize well to new scenes, producing high-fidelity 14--25 frame generations from a single seed image. Concurrent work on generation with motion guidance, including MotionCtrl~\cite{wang2024motionctrl} and CameraCtrl~\cite{he2025cameractrl}, showed that incorporating explicit control signals, such as camera trajectories or optical flow fields into the conditioning mechanism, substantially improves physical plausibility. Prior work has further demonstrated that rendering intermediate representations, such as skeleton keypoints \cite{chan2019everybody}, optical flow maps, or depth sketches, as image-space guidance effectively decouples the source of motion information from the visual domain \cite{zhang2023adding}.

Our work builds on this line of research by replacing camera or dense-flow conditioning with sparse object trajectory data via SG-I2V~\cite{namekata2024sg}, a more practically accessible signal in sensor-equipped monitoring systems.

\section{Methodology}
This work addresses the problem of synthesizing missing or dropped frames in top-down drone
video of Autonomous Surface Vehicles (ASVs) performing structured maritime maneuvers. Rather
than relying on generic video interpolation, we exploit the rich GPS telemetry available from the
vessels to condition a trajectory-guided image-to-video diffusion model. The resulting pipeline
converts real-world telemetry position logs into per-frame motion conditioning signals that are passed
directly to the SG-I2V diffusion model, enabling physically plausible video synthesis without any
domain-specific fine-tuning.
The complete pipeline, from raw drone footage and GPS logs through bounding-box initialization, GPS-to-image
mapping, conditioned video generation, and quantitative baseline comparison, is described in the
following subsections.

\subsection{Methods Overview}
\subsubsection{SG-I2V}
SG-I2V \cite{namekata2024sg} is a pre-trained state-of-the-art trajectory-conditioning framework built on top of Stable Video Diffusion (SVD)~\cite{blattmann2023stable}, a latent image-to-video diffusion model that generates temporally coherent clips from a single reference frame. SG-I2V accepts, alongside the reference image, a set of object descriptors, each a bounding box $\mathbf{b}_i = [x,y,w,h]$ and a trajectory array $\mathbf{T}_i \in \mathbb{R}^{N \times 2}$ of desired pixel-space positions across the $N$ frames to be generated. 

\subsubsection{RIFE}
Practical-RIFE~\cite{practicalrife} is a practical extension of RIFE (Real-Time Intermediate Flow Estimation)~\cite{huang2022real}, presented at ECCV 2022. Given two frames $I_0$ and $I_1$ bounding the missing segment, its 
IFNet directly estimates bidirectional intermediate flow fields 
$F_{0\to t}$ and $F_{1\to t}$ at a target time $t \in (0,1)$ through 
a coarse-to-fine refinement, trained via knowledge distillation. A lightweight fusion network then blends the warped frames while suppressing occlusion artifacts. 

\subsubsection{Optical Flow Extrapolation}
Missing frames are synthesized as a second baseline by warping the reference frame forward using dense optical flow estimated via the Farneback algorithm~\cite{farneback2003two}, as implemented in OpenCV. The method approximates each pixel's neighborhood with a polynomial expansion and tracks its deformation between frames, yielding a dense $H \times W \times 2$ displacement field used to warp the reference frame to each target timestep. In this work, RIFE and Optical Flow operate purely from the bounding reference frames, with no access to GPS or trajectory data. 
\subsection{Dataset Description}
The data consists of a continuous top-down footage of two small boats (vessels) recorded by a DJI drone along with GPS metadata. For this work, we utilize a clip of two seconds, where the drone footage is relatively close to the two vessels along the sea, extracted from the full 21 seconds video covering an overtaking scenario. Key video properties, original and post-processing, are listed in Table~\ref{tab:fps_res}.

\begin{table}[h]
    \caption{Resolutions and Frame Rates}
    \label{tab:fps_res}
    \centering
    \begin{tabular}{l|c|c}
       \hline
       & Resolution & Frame Rate \\
       \hline
       Original & 3840 × 2160 pixels & 30 FPS \\
       \hline
       Post-processing & 1024 × 576 pixels & 7 FPS \\
       \hline
    \end{tabular}
\end{table}

Both vessels broadcast telemetry position reports at approximately 1-second intervals throughout the recording. The full GPS log spans approximately 90 seconds while the drone footage covers 21 seconds, and the two share no common hardware clock. Temporal alignment was established by visually correlating vessel motion patterns in the video with corresponding speed and heading changes in the GPS log, yielding a fixed time offset of $t_{\text{log}} = t_{\text{video}} + 21$\,s. This alignment is inherently approximate and introduces a residual temporal uncertainty that contributes to the Trajectory Accuracy metric later discussed in Section~\ref{subsec:results}. After alignment, only the two-second sub-clip and its corresponding log entries are used for the subsequent experiments.

\begin{table}[h]
    \caption{Attributes in the ASV logs file}
    \label{tab:asv_logs}
    \centering
    \begin{tabular}{c|c|c}
    \hline
    ID & Colour & SOG range (kn) \\
    \hline
       99999 & Green& 0 - 3.5 \\
       \hline
       100000 & Yellow & 0 - 4.2 \\
       \hline
    \end{tabular}
\end{table}

Each log row carries: UTC timestamp, ID, longitude, latitude, Speed-Over-Ground (SOG, knots),
Course-Over-Ground (COG, degrees), magnetic heading, and a mission phase UUID.

Since drone telemetry was unavailable for this recording, the camera yaw (i.e. \ the compass direction corresponding to the top of the video frame) was estimated directly from the experimental data. Both ASVs were equipped with GPS receivers whose logs include the \textit{Course Over Ground} (COG), which is the direction of actual vessel movement derived from successive GPS fixes, expressed in degrees clockwise from North. At the starting timestamp, the green vessel (ID 99999) reported a COG of $89^\circ$ and the yellow vessel (ID 100000) reported a COG of approximately $86^\circ$, indicating that both were travelling nearly due East.

Inspecting the corresponding video frame, both vessels are observed moving toward the upper portion of the image, with their wake trailing toward the lower-left. This visual correspondence implies that the "up" direction of the image aligns closely with East on the map, placing the camera yaw in the vicinity of $90^\circ$.

In the absence of drone telemetry that would otherwise provide a precise
heading, a value of $\theta = 100^\circ$ was adopted as a conservative estimate, acknowledging that the drone may not have been oriented exactly
East and allowing for a small rotational offset from the direction of
vessel travel.

\begin{figure}[ht]
\vspace{1mm}
    \centering
    \includegraphics[width=0.85\columnwidth]{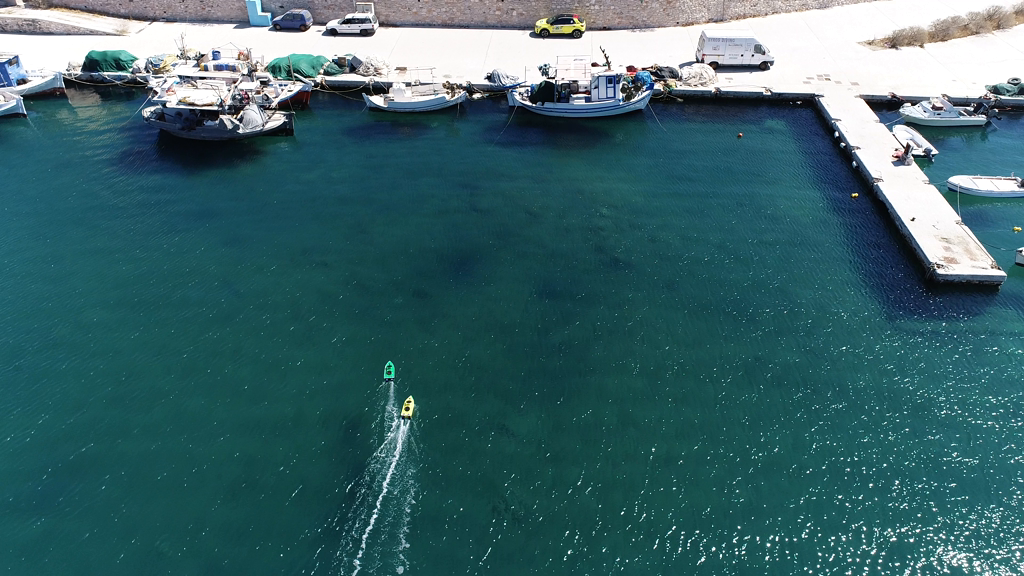}
    \caption{First frame of the video footage}
    \label{fig:first_frame}
\end{figure}

The first frame of the two second video can be seen in Figure~\ref{fig:first_frame}.

\subsection{Pipeline}
The proposed pipeline translates raw GPS telemetry and a single keyframe into a photorealistic video sequence. It proceeds through four stages: bounding-box     
initialization, GPS-to-pixel projection, trajectory-conditioned video generation,  
and quantitative evaluation.

\subsubsection{SG-I2V input construction}\label{sec:bbox}

\textbf{Bounding-box construction.}
Given the reference frame (first frame of the 2-second clip), the user clicks the center of each vessel in an interactive \texttt{matplotlib} window. Manual initialization is used here rather than automated detection, as the primary challenge in this setting is not detection but identity-consistent assignment of GPS trajectories to specific vessels. An automated detector such as YOLO could localize vessels, but without a reliable tracking and re-identification mechanism, correctly associating each detected bounding box with its corresponding GPS log entry, particularly when vessels are in close proximity or partially occluded, cannot be guaranteed.
Each click is therefore wrapped into a square bounding box, yielding a $[x, y, w, h]$ box centered on the clicked pixel. In addition to the two vessel boxes, four corner bounding boxes are placed at the image corners to encode global camera motion for SG-I2V conditioning.

\begin{figure}[ht]
\vspace{1mm}
    \centering
    \includegraphics[width=0.85\columnwidth]{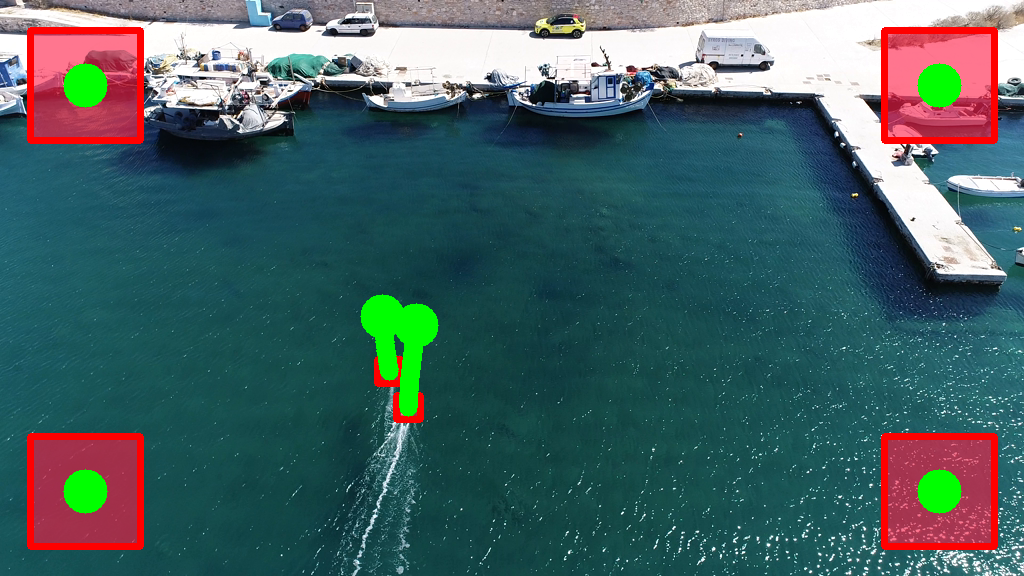}
    \caption{Conditioned input to SG-I2V showing the reference frame 
annotated with vessel bounding boxes and corner motion anchors.}
    \label{fig:cond_input}
\end{figure}

Since the drone remained almost stationary during the clip, each corner trajectory is held constant at its fixed center pixel for all $N = 14$ frames, signalling to SG-I2V that no global camera motion is present. Corner boxes have a size of $35$\,px, positioned $30$\,px inward from each image edge. The final conditioning payload for SG-I2V therefore consists of exactly six entries, green vessel, yellow vessel, top-left, top-right, bottom-left, bottom-right, each represented as a pair $\bigl(\mathbf{b}_i,\,\mathbf{T}_i\bigr)$ where $\mathbf{b}_i = [x, y, w, h]$ is the bounding box and $\mathbf{T}_i \in \mathbb{R}^{N \times 2}$ is the trajectory array with $N = 14$. The conditioned input to SG-I2V can be seen in Figure~\ref{fig:cond_input}.

\textbf{GPS-to-Pixel Mapping Function.}
The vessel trajectories are derived entirely from the ASV logs, without any optical tracking. The procedure follows three steps. 

\textit{Step 1 — Geographic-to-metric frame transform.}
Each GPS fix $(lon, lat)$ is converted to a local East-North metric offset relative to a 
reference origin $(\bar{lon}, \bar{lat})$, the mean position across all log entries, using 
the standard equirectangular approximation:
\begin{align}
    e &= (lon - \bar{lon}) \cdot M_{lon} \\
    n &= (lat - \bar{lat}) \cdot M_{lat}
\end{align}
where $M_{lat} = 111{,}320$\,m/deg is a standard geodetic constant representing the 
metres-per-degree along a meridian, and $M_{lon} = M_{lat} \cdot \cos(\bar{lat})$ 
accounts for the convergence of longitude degrees at the mean latitude of the scene.
The resulting East-North vector is then rotated by the estimated drone yaw 
$\theta = 100^\circ$, a value derived from visual inspection of the reference frame 
combined with the COG readings in the ASV logs, to align the metric frame with the 
image axes:
\begin{equation}
    \begin{pmatrix} f_x \\ f_y \end{pmatrix}
    =
    \begin{pmatrix} \cos\theta & -\sin\theta \\ \sin\theta & \cos\theta \end{pmatrix}
    \begin{pmatrix} e \\ n \end{pmatrix}
\end{equation}

\textit{Step 2 — Pixel-per-metre scale estimation.}
The image scale $s$ (px/m) is estimated directly from the reference frame without 
requiring drone altitude or camera intrinsics \cite{criminisi2000single}. At the anchor timestamp $t_0$, both 
vessels are projected into the metric frame via Step~1, yielding a metre-domain 
inter-vessel distance $d_m$. The corresponding pixel distance $d_p$ is computed from 
the two manually clicked bounding-box centres in the reference frame. The scale is 
therefore:
\begin{equation}
    s = \frac{d_p}{d_m} \quad \text{[px/m]}
\end{equation}
Note that $d_p$ depends on the manual click and is not derived from the CSV. For the 
recorded clip this yields $s = 28.3$\,px/m.

\textit{Step 3 — Per-frame pixel projection.}
For each of the $N = 14$ trajectory frames, the GPS log is linearly interpolated to 
the corresponding video timestamp $t_i = t_{\text{start}} + i\,/\,\text{fps}$. The 
interpolated position is converted to a metric offset $(\Delta f_x, \Delta f_y)$ 
relative to the vessel's anchor position at $t_0$ via Step~1, then projected into 
pixel space by applying the scale $s$ and offsetting from the manually clicked 
bounding-box centre $(c_x, c_y)$:
\begin{equation}
    p_i = \bigl(c_x + \Delta f_x \cdot s,\;\; c_y - \Delta f_y \cdot s\bigr)
\end{equation}

The sign inversion on $\Delta f_y$ accounts for the image coordinate system, in which 
the $y$-axis increases downward.

\subsection{Video Generation}

After constructing the SG-I2V conditioning payload from the previous steps, the model 
is run in inference mode to generate the 14 subsequent frames. Since the vessels are 
small relative to the frame and the scene consists of low-texture open water, the 
domain is significantly out of distribution with respect to the large-scale video 
corpora on which the SVD backbone was pre-trained. No domain-specific fine-tuning is 
applied. Instead, a systematic 
ablation study over key hyperparameters governing trajectory guidance 
strength, specifically the number of latent optimisation iterations, 
the heatmap spatial spread, and the FFT frequency blend ratio, was 
conducted across multiple experimental runs. The configuration yielding the 
best balance of visual quality and temporal coherence upon inspection 
was selected for the final evaluation.

\section{Evaluation}\label{sec:evaluation}

\subsection{Evaluation Metrics}
All methods are evaluated against the corresponding ground-truth video segment 
(frames 2--15 of the clip) across four categories of metrics.

\textbf{Perceptual quality.}
Learned Perceptual Image Patch Similarity (LPIPS)~\cite{zhang2018unreasonable} compares deep 
feature representations rather than raw pixels, using an AlexNet backbone. Frames are 
resized to $256 \times 256$ and normalised to $[-1, 1]$ prior to scoring. Lower values 
indicate higher perceptual similarity.

\textbf{Temporal consistency.} Temporal smoothness measures the mean Farneb\"{a}ck flow magnitude 
within each video independently, reported in px/frame. Values close to the ground-truth 
reference indicate that the generated video reproduces realistic amounts of scene motion; 
large deviations indicate over- or under-estimated motion.

\textbf{No-reference quality.}
BRISQUE (Blind/Referenceless Image Spatial Quality Evaluator)~\cite{mittal2012no} 
estimates perceptual image naturalness without requiring ground-truth reference frames, 
computed using the \texttt{piq} library. Lower scores indicate more natural-looking 
frames. BRISQUE is reported separately for generated and ground-truth frames to allow 
direct comparison of absolute naturalness.

\textbf{Trajectory accuracy.}
Vessel trajectory accuracy is evaluated by tracking the bounding-box centres of both 
vessels forward through each method's generated frames using Lucas--Kanade (LK) sparse optical 
flow~\cite{lucas1981iterative} with a $21 \times 21$ pixel search window and a three-level 
Gaussian pyramid. The tracked pixel positions are compared against GPS-projected 
ground-truth positions derived from the ASV logs using the same coordinate mapping 
function described in Section~\ref{sec:bbox}. Trajectory error is reported in pixels 
separately for the green vessel (ID 99999) and yellow vessel (ID 100000), as well 
as their mean. This metric assumes reliable LK feature correspondence throughout the 
generated sequence, a condition that holds when frames maintain sufficient visual 
coherence across all 14 frames, an assumption that may not hold equally across all 
methods, as discussed in Section~\ref{subsec:results}.

\section{Results}
\label{subsec:results}

Table~\ref{tab:quantitative} reports all quantitative metrics across all three methods and the ground-truth reference.

\begin{table}[h]
    \caption{Quantitative comparison across all methods. $^\dagger$BRISQUE 
    is a no-reference metric (lower = more natural); ground truth score (23.64) 
    serves as a naturalness reference rather than a target to minimise.}
    \label{tab:quantitative}
    \centering
    \small
    \begin{tabularx}{\linewidth}{|l|X|X|X|X|}
        \hline
        \textbf{Method} & \textbf{LPIPS} (lower) & \textbf{Temp.\ smooth} & \textbf{BRISQUE} $^\dagger$ (lower) & \textbf{Traj err px} (lower) \\
        \hline
        SG-I2V     & 0.149          & \textbf{1.14}  & \textbf{25.52} & \textbf{9.31}  \\
        \hline
        Opt.\ Flow & \textbf{0.110} & 0.39           & 17.47          & 27.71          \\
        \hline
        RIFE       & 0.130          & 0.59           & 42.46          & 34.75          \\
        \hline
        Ground truth (GT) ref.\   & ---            & 1.42           & 23.64          & 28.70          \\
        \hline
    \end{tabularx}
\end{table}

Optical flow extrapolation scores best on perceptual similarity (LPIPS 0.110), 
though this reflects the preservation of original image statistics through direct 
frame warping rather than genuine perceptual quality. RIFE, benefiting from 
bidirectional interpolation between known endpoint frames, an advantage 
unavailable in genuine missing-frame scenarios where future frames do not exist, produces smooth but over-averaged outputs, as evidenced by its BRISQUE score 
of 42.46, the worst among all methods. SG-I2V produces the most realistic motion 
magnitude (temporal smoothness 1.14, closest to ground truth 1.42), the most naturally 
appearing generated frames among synthesis methods (BRISQUE 25.52, closest to ground truth 
23.64), and the strongest trajectory adherence (9.31px, well below the GPS mapping 
noise floor of 28.70px established by the ground-truth video itself).

\begin{figure*}[t]
\vspace{3mm}
    \centering
    \includegraphics[width=1.0\textwidth, height=0.275\textheight, keepaspectratio=false]{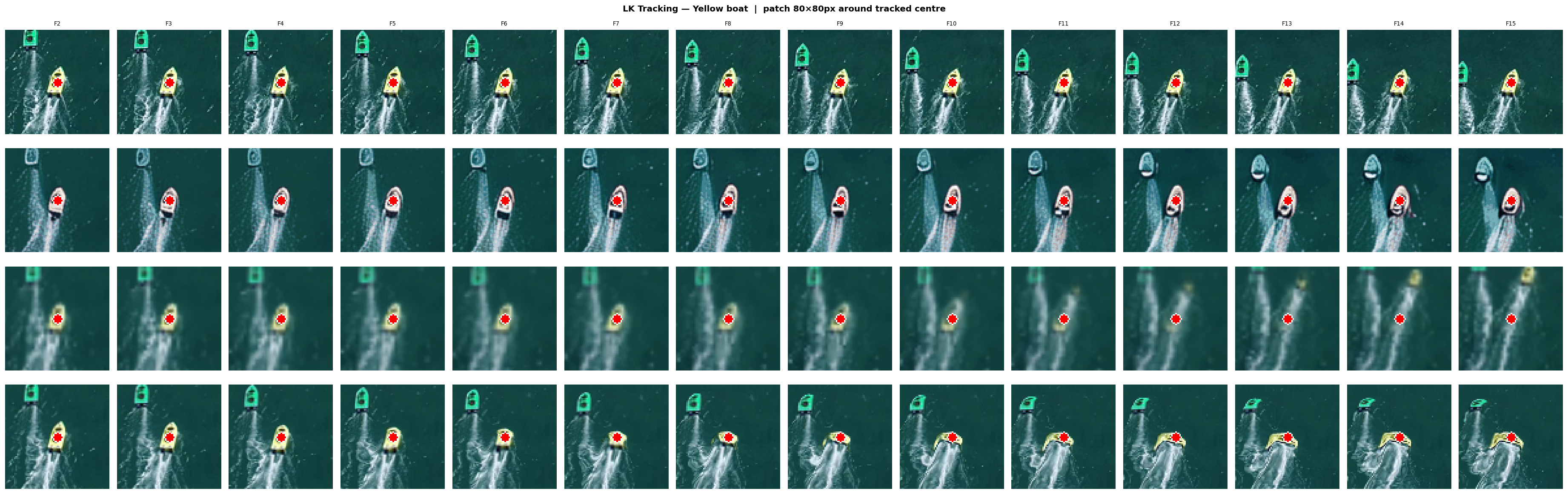}
    \caption{LK tracking patches for the yellow vessel across all 14 generated frames. 
    Each row corresponds to a method (Ground Truth, SG-I2V, RIFE, Optical Flow) and 
    each column to a frame (F2--F15). The red dot indicates the LK-tracked vessel 
    centre.}
    \label{fig:patches}
\end{figure*}

Regarding trajectory accuracy, the metric is most directly interpretable for 
SG-I2V, where LK tracking remains stable across all 14 frames as confirmed by 
visual inspection of the tracking patches (Figure~\ref{fig:patches}). For RIFE 
and Optical Flow, progressive frame degradation, over-smoothing artifacts and 
accumulated warp blur respectively, reduces tracking reliability in later 
frames, potentially conflating positional deviation with tracking failure and 
inflating reported errors. Furthermore, the manual temporal alignment between the 
drone footage and ASV logs, necessary in the absence of a shared hardware clock, 
introduces a residual positional uncertainty evidenced by the ground-truth video 
itself scoring 28.70px. Consequently, the trajectory metric serves primarily as 
evidence of SG-I2V's GPS conditioning effectiveness: the model successfully 
incorporates the GPS-derived motion cues into the generated frames, producing 
vessel positions that closely adhere to the projected trajectories. The comparison 
between methods on perceptual and temporal metrics, which are independent of the 
GPS coordinate system, provides a more reliable basis for evaluating generation 
quality overall.

\section{Conclusion}

This paper presented a pipeline for reconstructing missing frames in top-down maritime drone video by conditioning a pre-trained image-to-video diffusion model on GPS-derived vessel trajectories, requiring no domain-specific fine-tuning. A GPS-to-Pixel mapping function converts sparse telemetry position 
logs into spatial motion cues consumable by SG-I2V, bridging the gap between physical sensor 
data and generative video synthesis.

SG-I2V produces the most naturally appearing frames (BRISQUE 25.52, closest to ground truth 23.64), the most realistic motion magnitude (temporal smoothness 1.14, closest to ground truth 1.42), and the strongest GPS trajectory adherence (9.31px vs. 28.70px for ground truth), confirming that trajectory-guided diffusion synthesis is a viable approach to maritime video reconstruction under challenging low-texture, small-object conditions. The trajectory evaluation further confirms that GPS-derived conditioning effectively steers vessel positions toward physically plausible locations, with the ground truth baseline error of 28.70px attributable to the approximate manual temporal alignment between footage and GPS logs rather than to any generation artifact.

The primary limitation of this work is the evaluation on a single 
two second clip, which, while sufficient to demonstrate the viability 
of the proposed approach, limits the generalizability of the quantitative 
findings. Future work would extend the pipeline to longer sequences 
and more complex multi-vessel scenarios, explore automatic 
synchronization methods between video footage and auxiliary sensor 
logs, and replace the manual bounding box initialization with an automated detection and tracking pipeline, for example using YOLO combined with a re-identification module, toward a fully automated system, provided that consistent vessel-to-log assignment can be maintained throughout the sequence. Further work would also investigate fine-tuning strategies to better adapt the SVD backbone to smaller objects. A particularly promising 
direction is the incorporation SOG data, 
already available in the ASV logs, into the trajectory conditioning 
payload. Currently, projected arrow overlays encode the direction of the vessel,
but not the magnitude of displacement scaled by speed. Explicitly 
scaling each vessel's trajectory length by its SOG relative to the 
other vessels would allow the conditioning signal to encode physically 
accurate inter-vessel speed differences.
\section*{Acknowledgment}
This work was supported by the MUSIT Project through the European Union’s Horizon Europe Framework Programme (HORIZON), under Marie Sklodowska-Curie grant agreement no. 101182585. The work only reflects the authors’ views; the EU Agency is not responsible for any use of the information it contains.

\bibliographystyle{IEEEtran}
\bibliography{references}

\end{document}